%% file: main_RAL.tex
\documentclass[letterpaper, 10 pt, conference]{ieeeconf}  % Comment this line out if you need a4paper

\IEEEoverridecommandlockouts                              % This command is only needed if    
\usepackage{mathptmx} % assumes new font selection scheme installed
\usepackage{times} % assumes new font selection scheme installed
\usepackage{bbding}
\usepackage{graphicx}
\usepackage{float}
\usepackage{diagbox}
\usepackage{arydshln} 
\usepackage{bbm}
\usepackage{wrapfig}
\usepackage{mathrsfs}
\usepackage{listings} 
\usepackage{physics}

\usepackage{graphics} % for pdf, bitmapped graphics files
\usepackage{float}
\usepackage{multirow}
\usepackage{amsmath}
\usepackage{bbm}
\usepackage{amssymb}
\usepackage{mathtools}
\usepackage[utf8]{inputenc} % allow utf-8 input
\usepackage[T1]{fontenc}    % use 8-bit T1 fonts
\usepackage{hyperref}       % hyperlinks
\usepackage{cleveref}
\usepackage{url}            % simple URL typesetting
\usepackage{booktabs}       % professional-quality tables
\usepackage{amsfonts}       % blackboard math symbols
\usepackage{nicefrac}       % compact symbols for 1/2, etc.
\usepackage{microtype}      % microtypography
\usepackage{xcolor}         % colors
\usepackage{algorithm}
\usepackage{algorithmicx}
\usepackage{algorithm} % Required for insertion of code
\usepackage{algpseudocode}
\usepackage{fancyvrb}
\usepackage{fvextra}
\usepackage{csquotes}

\usepackage{tcolorbox}
\usepackage{multicol}
\tcbuselibrary{skins}

\title{\LARGE \bf
LVDiffusor: Distilling Functional Rearrangement Priors from\\ Large Models into Diffusor
}

\author{
Yiming Zeng*, Mingdong Wu*, Long Yang, Jiyao Zhang, Hao Ding, Hui Cheng, Hao Dong
\thanks{*Equal contribution.}
\thanks{Yiming Zeng, Hao Ding, and Hui Cheng are with the School of Computer Science and Engineering, Sun Yat-Sen University.
Mingdong Wu, Jiyao Zhang, Long Yang and Hao Dong are with Hyperplane Lab, School of CS, Peking University and National Key Laboratory for Multimedia Information Processing.
Jiyao Zhang is also with the Beijing Academy of Artificial Intelligence (BAAI).
}
\thanks{%\textsuperscript{\Letter} 
Corresponding to hao.dong@pku.edu.cn and chengh9@mail.sysu.edu.cn}
}

\begin{document}
 
% notations

\input{notations}

\maketitle

\begin{abstract}
Object rearrangement, a fundamental challenge in robotics, demands versatile strategies to handle diverse objects, configurations, and functional needs. To achieve this, the AI robot needs to learn functional rearrangement priors to specify precise goals that meet the functional requirements. Previous methods typically learn such priors from either laborious human annotations or manually designed heuristics, which limits scalability and generalization.
In this work, we propose a novel approach that leverages large models to distill functional rearrangement priors. 
Specifically, our approach collects diverse arrangement examples using both LLMs and VLMs and then distills the examples into a diffusion model. 
During test time, the learned diffusion model is conditioned on the initial configuration and guides the positioning of objects to meet functional requirements. 
In this manner, we create a handshaking point that combines the strengths of conditional generative models and large models. 
Extensive experiments on multiple domains, including real-world scenarios, demonstrate the effectiveness of our approach in generating compatible goals for object rearrangement tasks, significantly outperforming baseline methods. 
Our real-world results can be seen on \url{https://sites.google.com/view/lvdiffusion}.
\end{abstract}

\vspace{-5pt}
\section{INTRODUCTION}
% \vspace{-5pt}

%%%%%%%%%%%%%%%%%%%%%%%%%%%%%%%%%%%%%%%%%%%%%%%%%%%%%%%%%%%%%%%%%%%%%%%%%%%%
% Problem and Difficulties
%%%%%%%%%%%%%%%%%%%%%%%%%%%%%%%%%%%%%%%%%%%%%%%%%%%%%%%%%%%%%%%%%%%%%%%%%%%%
Object rearrangement~\cite{rearrange} is a fundamental challenge in robotics that is widely encountered in our daily lives, such as when organizing a cluttered writing desk, reconfiguring furniture in the bedroom, or setting up a dining table for a left-handed user. 
This task requires the AI robot to specify precise goals, including object locations, and then rearrange the objects to achieve those goals.
To accomplish this, the AI robot needs to learn \textit{functional rearrangement priors}, \ie, how to position the objects to fulfill functional requirements. 
These priors should be capable of handling diverse objects, configurations, and functional needs. 
This presents a significant challenge for robotics, as manually designing a reward or goal function is difficult~\cite{wu2022targf}.

Previous works~\cite{shen2020structformer, liu2022structdiffusion, simeonov2023shelving, neatnet, wu2022targf} typically learn such priors from a dataset either manually designed by humans or synthesized using heuristic rules. However, the former approach requires laborious human annotations, limiting its \textbf{scalability}. The latter approach depends on hard-coded heuristics, making it difficult to \textbf{generalize} to diverse configurations.
% On the other hand, DALL-E-Bot~\cite{dallebot} proposes an alternative direction that leverages an internet-scale pre-trained diffusion model to generate the arrangement goal according to the initial state. 
% However, this approach suffers from a significant \textit{compatibility issue}: there is no guarantee that the generated goal will be \textbf{compatible} with the ground-truth configuration. Additionally, this method is time-consuming as it requires multiple inferences from a large model.
On the other hand, DALL-E-Bot~\cite{dallebot} proposes an alternative direction that leverages an internet-scale pre-trained diffusion model to generate the arrangement goal according to the initial state. 
Despite its success, this approach faces an inherent \textit{compatibility issue} brought by the VLM: there is no guarantee that the generated goal will be \textbf{compatible} with the ground-truth configuration since the VLM-generated image may not align with the language prompt~(\eg, object number and category). As a result, this method tends to be time-consuming as it necessitates multiple inferences from a large model to pass the filter.
Given these, we pose the following question:

\begin{displayquote}
    \textit{How to learn \textbf{generalizable} functional rearrangement priors that can generate \textbf{compatible} goals for diverse configurations, in a \textbf{scalable} manner?}
\end{displayquote}

%%%%%%%%%%%%%%%%%%%%%%%%%% Teaser %%%%%%%%%%%%%%%%%%%%%%%%%%%%%%%%%%
\begin{figure}[t]
\begin{center}
\includegraphics[width=\linewidth]{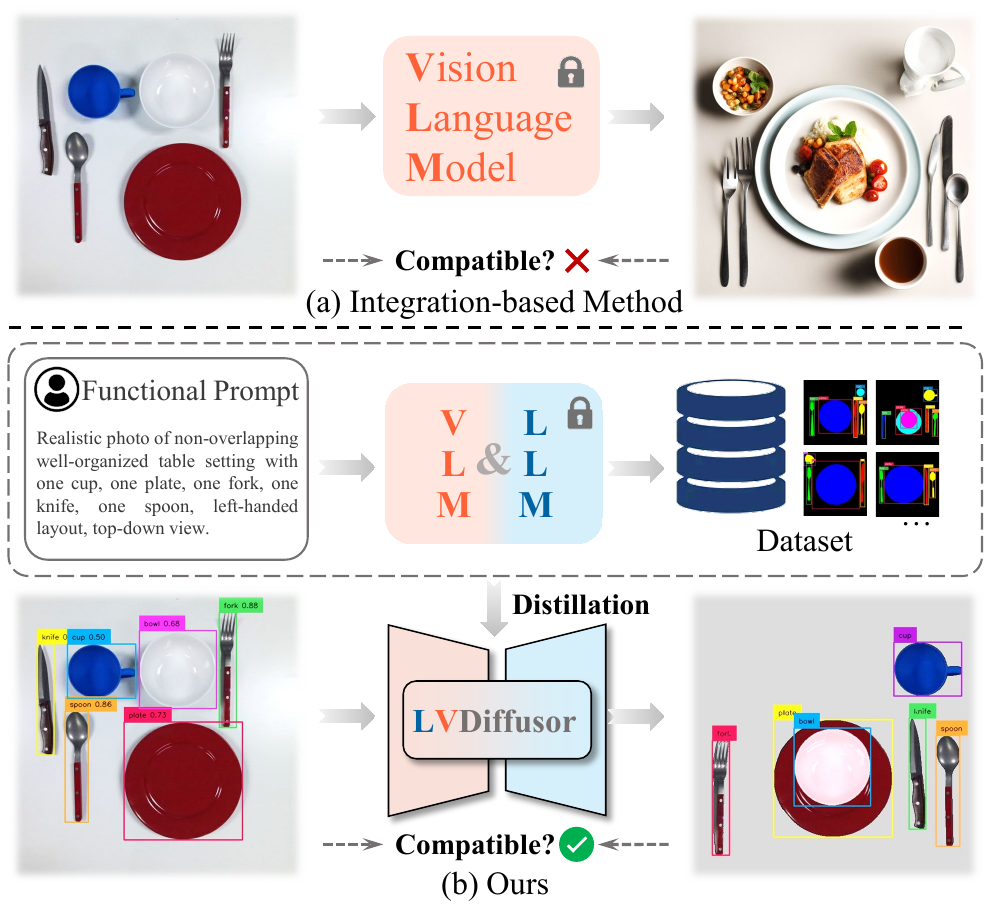}
\end{center}
\vspace{-15pt}
\caption{\textbf{(a):} Integrating a large model into the rearrangement pipeline may lead to \textit{compatibility issues}.
\textbf{(b):} Differently, we \textit{distill} a conditional generative model from the large models, which helps alleviate this issue.}
\label{fig:teasor}
\vspace{-20pt}
\end{figure}
%%%%%%%%%%%%%%%%%%%%%%%%% Teaser %%%%%%%%%%%%%%%%%%%%%%%%%%%%%%%%%%%%%%%%%%%%

%%%%%%%%%%%%%%%%%%%%%%%%%%%%%%%%%%%%%%%%%%%%%%%%%%%%%%%%%%%%%%%%%%%%%%%%%%%
% Method to address difficulties
%%%%%%%%%%%%%%%%%%%%%%%%%%%%%%%%%%%%%%%%%%%%%%%%%%%%%%%%%%%%%%%%%%%%%%%%%%%

As shown in Figure~\ref{fig:teasor}, our key idea is to distill functional rearrangement priors from large models into compact representations. 
Firstly, we prompt a large Visual-Language-Model (VLM), such as StableDiffusion~\cite{rombach2022high}, to generate a dataset filled with arrangement examples (\ie, goals) that satisfy the functional needs. 
Similar with~\cite{liu2022structdiffusion}, we can then train a conditional generative model~(\eg, diffusion model~\cite{SDEScoreMatching}) on these arrangement examples to model the distilled functional rearrangement priors. 
In this manner, we create a handshaking point that combines the strengths of the conditional generative model and large models. 
On one hand, prompting large models allows us to collect arrangement examples across a wide range of diverse configurations, facilitating generalization in a scalable manner. 
On the other hand, distilling the gathered data into a conditional generative model enables us to generate feasible goals that are compatible with the initial conditions.

Unfortunately, as also observed by~\cite{lee2023aligning, black2023training}, the output of the VLM cannot be guaranteed to be consistent with the prompt input, which may lead us to distill incorrect knowledge from large models. 
Specifically, the generated results may deviate from the specified number and types of objects in the prompt, and may not perfectly align with the functional requirements outlined in the prompt.
For instance, when prompted with ``a well-organized arrangement of a fork, two bowls, and a plate on a dinner table'' the VLM may generate an arrangement with two randomly placed bowls~(\emph{i.e.}, \textit{violating functional requirement}) multiple forks~(\emph{i.e.}, \textit{different number}), and a mug~(\emph{i.e.}, \textit{an unexpected object type}).

To alleviate this issue, we integrate a Large Language Model after prompting the VLM to assist in correcting the generated examples.
We initially instruct the VLM with an original prompt, such as ``Realistic photo of the non-overlapping, well-organized table setting with one cup, one plate, one fork, one knife, one spoon, left-handed layout, top-down'', and extract object states (\emph{e.g.}, bounding boxes) through an off-the-shelf perception module (\eg, DINO~\cite{liu2023grounding}).
Afterward, we input these object states and prompt the Large Language Model (LLM) using a Chain-of-Thoughts approach~\cite{wei2022chain} to fine-tune their positions, ensuring human-like layouts that better align with the functional requirements.

%%%%%%%%%%%%%%%%%%%%%%%%%%%%%%%%%%%%%%%%%%%%%%%%%%%%%%%%%%%%%%%%%%%%%%%%%%%
% Results
%%%%%%%%%%%%%%%%%%%%%%%%%%%%%%%%%%%%%%%%%%%%%%%%%%%%%%%%%%%%%%%%%%%%%%%%%%%

We conduct experiments in multiple scenarios and functionalities to demonstrate the effectiveness of our approach in learning functional rearrangement priors and deploying them in real-world rearrangements. 
Extensive results and analysis showcase that our method significantly outperforms the baseline in generating compatible rearrangement goals for a large variety of configurations.
Ablation studies further suggest that both LLM and VLM play indispensable roles in distilling the functional rearrangement priors.

%%%%%%%%%%%%%%%%%%%%%%%%%%%%%%%%%%%%%%%%%%%%%%%%%%%%%%%%%%%%%%%%%%%%%%%%%%%
% Conclusions
%%%%%%%%%%%%%%%%%%%%%%%%%%%%%%%%%%%%%%%%%%%%%%%%%%%%%%%%%%%%%%%%%%%%%%%%%%%

In summary, our contributions are summarized as follows:
\begin{itemize}
    \item 
    We introduce a novel framework that trains a diffusion model to distill functional rearrangement priors from both LLM and VLM for object rearrangement.
    % \textcolor{blue}{This distillation approach greatly generates compatible goal layouts that align with various initial settings. }
    \item 
    We propose leveraging the LLM, such as GPT4, to help alleviate the misalignment between the generated results and the VLM prompt.
    \item 
    We conduct extensive experiments to demonstrate the effectiveness of our approach in generating compatible goals and real-world deployment.
\end{itemize}

\vspace{-5pt}
\section{RELATED WORK}
\vspace{-5pt}

\subsection{Object Rearrangement with Functional Requirements}
The object rearrangement is a fundamental challenge~\cite{rearrange} and a long-studied problem\cite{abdo2015robot, schuster2012learning, abdo2016organizing} in robotics and the graphics community~\cite{fisher2012example, MakeItHome2011, RJMCMC2012, SceneNet2016, songchun2018, wei2023lego}.
One of the keys to tackling object rearrangement is goal specification, \ie, how to specify precise rearrangement goals to meet the functional requirements.
Early works typically focus on manually designing rules/ energy functions to find a goal~\cite{abdo2015robot, schuster2012learning, abdo2016organizing} or synthesizing a scene configuration~\cite{MakeItHome2011, RJMCMC2012, SceneNet2016, songchun2018} that satisfies the preference of the user. 

Some recent works learn such priors by training a conditional generative model from a dataset either manually designed
by humans or synthesized using heuristic rules. 
Neatnet~\cite{neatnet} tries to learn a GNN from a human-collected dataset to output an arrangement tailored to human preferences. 
TarGF~\cite{wu2022targf} and LEGO-NET~\cite{wei2023lego} learn gradient fields that provide guiding directions to rearrange objects, via Denoising-Score-Matching~\cite{denosingScoreMatching} from the indoor-scene dataset designed by artists.
StructDiffusion~\cite{liu2022structdiffusion}, StructFormer~\cite{shen2020structformer} and RPDiff~\cite{simeonov2023shelving} learn a conditional generative model from arrangement examples generated in simulation using a handcrafted function.
However, these methods either depend on hard-coded heuristics that make them difficult to generalize to diverse configurations or require laborious human annotations that limit the scalability.

Another stream of research attempts to leverage the Large Language Model~(LLM) or large Visual Language Model~(VLM) for goal specification. \cite{ECCV22HouseKeep, ECCV22TIDEE} notice the necessity of automatic goal inference for tying rooms and exploit the commonsense kn owledge from LLM or memex graph to infer rearrangements goals when the goal is unspecified.
TidyBot~\cite{wu2023tidybot} also leverages an LLM to summarize the rearrangement preference from a few examples provided by the user.
However, it is difficult for LLM to specify the coordinate-level goal,
The most recent work, DALL-E-Bot~\cite{dallebot}, integrates a large VLM, such as Dall-E 2~\cite{ramesh2022hierarchical} to generate the arrangement goal according to the initial state. However, there is no guarantee that the VLM-generated goal will be compatible with the ground-truth configuration.

Differing from those studies, our approach leverages the LLM and VLM to collaboratively collect arrangement examples, enabling generalizable and scalable data collection, and distills them into a conditional generative model.

\begin{figure*}[t]
\begin{center}
\includegraphics[width=\linewidth]{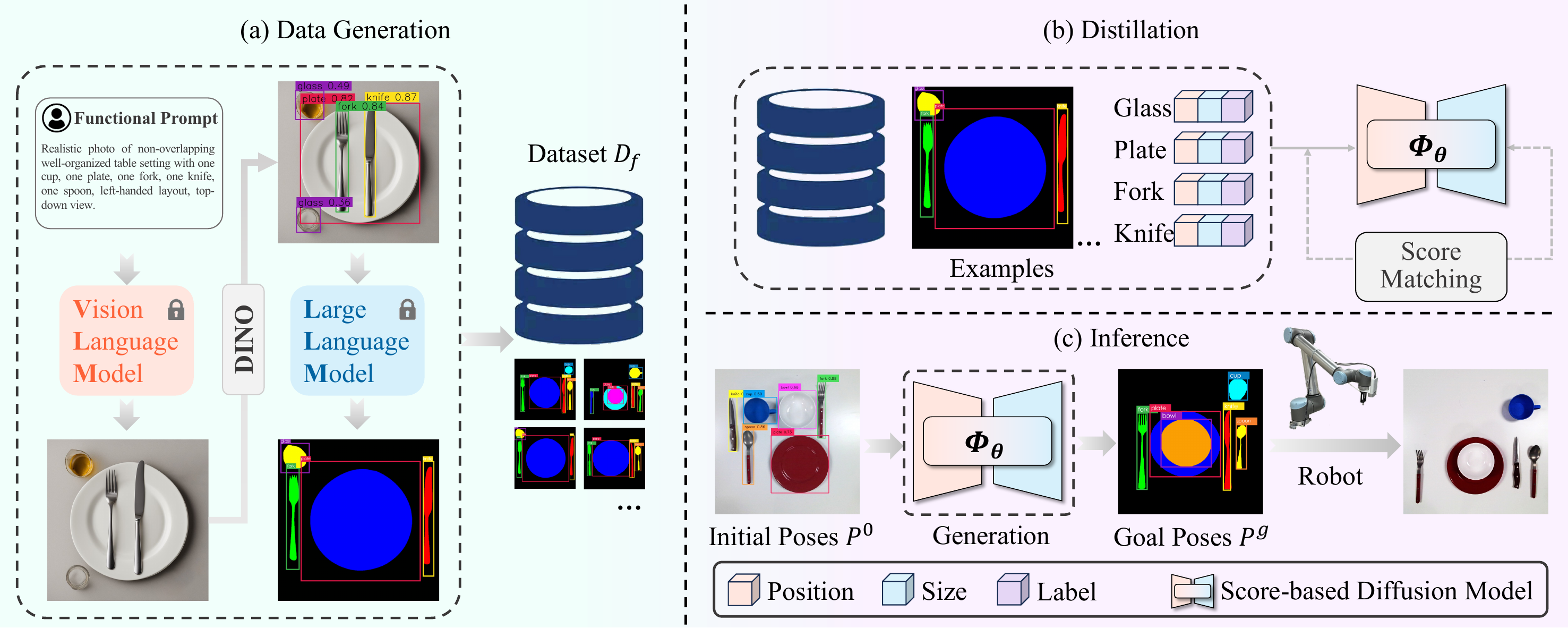}
\end{center}
\vspace{-10pt}
\caption{
\textbf{(a) Data Generation:} We construct an autonomous data collection pipeline to obtain arrangement examples, denoted as $\data = \{(\poses^i, \conds^i)\}_{i=1}^K$, in two stages, collaboratively using an LLM and a VLM. 
First, we generate initial arrangement examples, $\{(\hat{\poses^i}, \hat{\conds^i})\}_{i=1}^K$, by prompting the VLM and extracting object positions via GroundingDino. Then, we refine these examples using the LLM to obtain the final dataset, $\data$.
\textbf{(b) Distillation:} The collected dataset is distilled into a score-based diffusion model, denoted as $\score$, using a score-matching objective.
\textbf{(c) Inference:} During test time, we generate goal positions, $\goalposes$, using the learned diffusion model and rearrange objects from the initial positions, $\initposes$, to the goal positions, $\goalposes$.
}
\label{fig:pipeline}
% \vspace{-15pt}
\end{figure*}

\vspace{-5pt}
\subsection{Leveraging Large Models for Robot Learning}
Research on large models, typically Large Language Models~(LLMs)~\cite{OpenAI2023GPT4TR} and Visual Language Models~(VLMs)~\cite{rombach2022high, radford2021learning, palm-e, ramesh2022hierarchical}, has grown rapidly in recent years.
Inspired by this, RT-1~\cite{brohan2022rt} and RT-2~\cite{RT-2} explore the training of large models using large-scale demonstrations for robotic manipulation tasks. However, this necessitates a challenging data collection process that takes years to complete.

As a result, the robotics community has started to integrate the pre-trained large models into the robot learning workflow~\cite{huang2023visual, zhou2023navgpt, huang2023voxposer, lin2023text2motion, liang2023code, ahn2022can, dallebot, ECCV22HouseKeep, ECCV22TIDEE, wu2023tidybot}.
In navigation, VLMaps~\cite{huang2023visual} and NavGPT~\cite{zhou2023navgpt} leverage the LLM to translate natural language instructions into explicit goals or actions. 
In manipulation, SayCan~\cite{ahn2022can} leverages VLM to generate proposals for a given demand.
Code-as-Policies~\cite{liang2023code} and Text2motion~\cite{lin2023text2motion} leverage LLM to generate high-level plans for long-horizon tasks.
VoxPoser~\cite{huang2023voxposer} employs VLM and LLM collaboratively to synthesize robot trajectories.
In rearrangement, DalleBot~\cite{dallebot}, TidyBot~\cite{wu2023tidybot}, \cite{ECCV22HouseKeep} and ~\cite{ECCV22TIDEE} leverages VLM or LLM for goal specification.

Different from integration-based methods, we explore a distillation-based approach that extracts the functional rearrangement priors from the large models.
A concurrent work~\cite{ha2023scaling} also distilled robot skills from LLM. Differently, we simultaneously distilled knowledge from both LLM and VLM, rather than just VLM, and we distilled more fine-grained (i.e., coordinate-level) functional rearrangement priors.

\section{METHOD}
\label{sec:method}
% \vspace{-5pt}

\textbf{Task Description:}
We aim to rearrange objects to meet functional requirements on a 2D planar surface, such as a dinner table, etc. In formal terms, we denote the functional requirements, such as "set up the dinner table for a left-handed person," as $\func$. The AI robot is provided with a visual observation $\obs \in \R^{3\times H \times W}$ of the scene from a top-down view. Assuming there are $N$ objects denoted as $\objs = [\obj_1, \obj_2, ... \obj_N]$, the objective of the AI robot is to specify a set of target poses $\goalposes = [\goalpose_1, \goalpose_2, ... \goalpose_N]$ and move the objects from their initial poses $\initposes = [\initpose_1, \initpose_2, ... \initpose_N]$ to target poses $\goalposes$.

Ideally, both the initial and goal poses should include 2D positions and 1D rotations, i.e., $\initpose_i, \goalpose_i \in \text{SE}(2)$. However, in this work, we only consider the 2D position, i.e., $\initpose_i, \goalpose_i \in \R^2$, since the existing RGB-based 2D pose estimation does not support a sufficient number of object categories.

\textbf{Overview:}
Our goal is to train a conditional generative model capable of generating rearrangement goals compatible with the initial object conditions. 
Initially, we collect a dataset (Sec.~\ref{sec:dataset}) using a Language Model (LLM) and a Vision Language Model (VLM), denoted as $\data = \{(\poses^i, \conds^i) \sim \dist(\poses, \conds)\}_{i=1}^K$, where $\dist$ represents the data distribution of $\data$. Here, $\conds = [\cond_1, \cond_2, \ldots, \cond_N]$, with $\cond_i = [\size_i, \cate_i]$, describing the condition of the $i$-th object $\obj_i$. Specifically, $\size_i \in \mathbb{R}^2$ denotes the object's 2-D axis-aligned bounding box dimensions, and $\cate_i \in \mathbb{R}^1$ is its category label.
Then, we train a score-based diffusion model (Sec.~\ref{sec:diffusion}), referred to as $\score$, to model the conditional distribution $\dist(\poses|\conds)$ using the distilled dataset.

During inference (Sec.~\ref{sec:rearrange}), we start by computing the initial object conditions $\conds$ from the visual observation $\obs$. 
Next, we generate a feasible goal $\goalposes$ conditioned on the initial object configurations $\conds$ using the trained diffusion model $\score$, and finally, we create a plan to achieve this goal.

\subsection{Collaboratively Collecting Data through LLM and VLM}
\label{sec:dataset}
To extract functional rearrangement prior knowledge from large models, we construct an autonomous data collection pipeline to obtain the arrangement examples $\data = \{(\poses^i, \conds^i) \sim \dist(\poses, \conds)\}_{i=1}^K$. 
This approach allows us to generate a large number of arrangement examples without the need for human labor across various configurations. 
To achieve this, we sample arrangement examples in two stages, collaboratively using LLM and VLM: we first generate initial arrangement examples $\{(\hat{\poses^i}, \hat{\conds^i})\}_{i=1}^K$ using VLM and then refine these examples using LLM to obtain the final examples $\data$.

At the first stage, for each instance, we initially prompt the VLM~(\ie, StableDiffusion XL~\cite{rombach2022high}) to get an image. The prompt is constructed as follows: 
\begin{displayquote}
    \textit{Realistic photo of <adjective> <setting description> with <num1> <object1>, <num2> <object2>, .., <functional layout>, <view point>.}
\end{displayquote}
Subsequently, we extract initial objects' positions ${\hat{\poses}}$ and conditions $\hat{{\conds}}$ from the image using a large vision model for object detection, \ie, GroundingDino~\cite{liu2023grounding}, for further refinement.
% Therefore, we leverage GroundingDino to detect the category, size, and position of each object within the image for further refinement.

\begin{figure}[t]
\begin{center}
\includegraphics[width=0.95\linewidth]{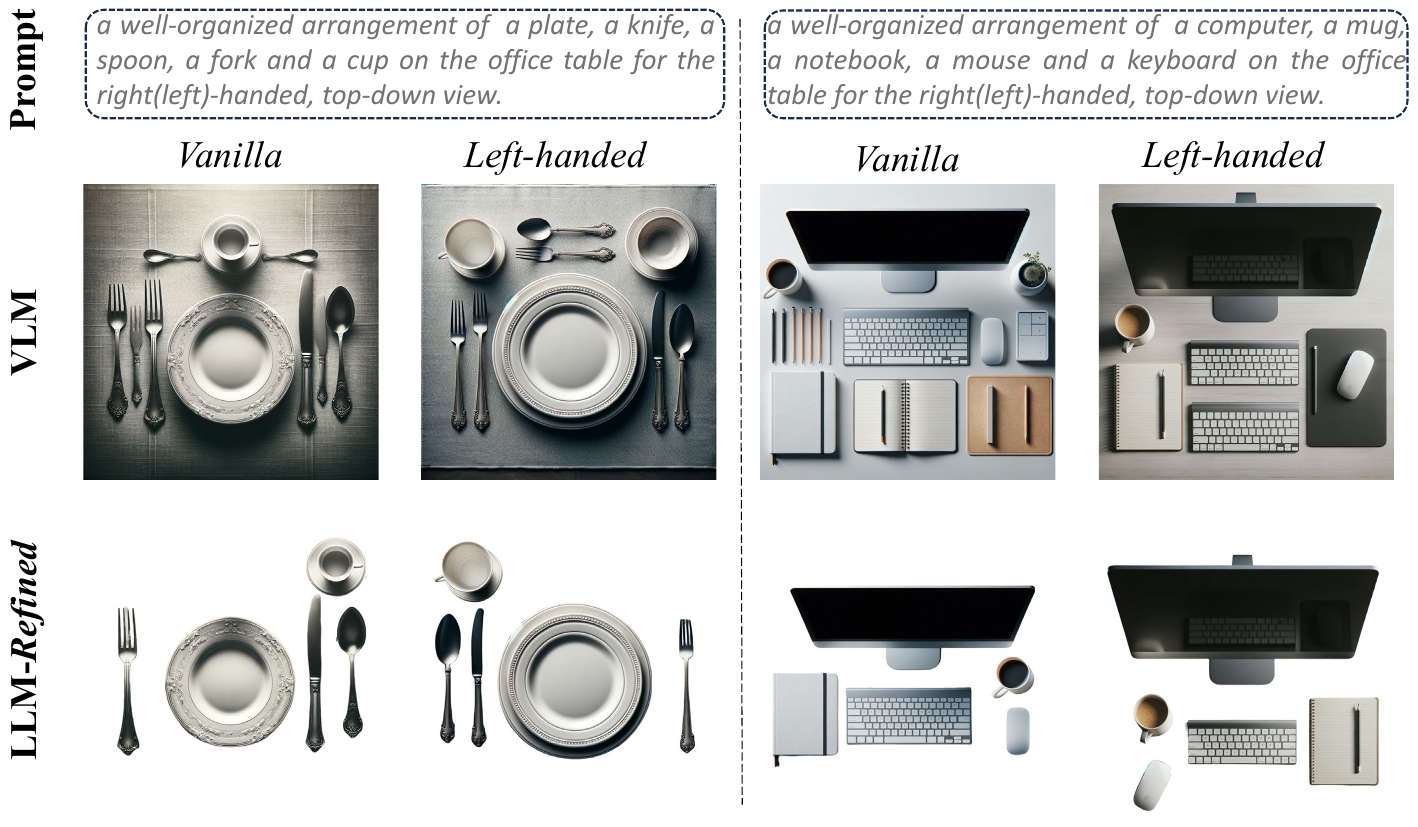}
\end{center}
\vspace{-10pt}
\caption{\textbf{Incompatible issue of VLM-generated layouts.} We leverage LLM to remove the redundant items and reposition the remaining ones to align with corresponding functional requirements. To visualize `LLM-Refined', we cropped the mask of each object from the results generated by VLM, and then moved the object mask into the LLM refined bounding box.}
\label{fig:incompatibility}
\vspace{-10pt}
\end{figure}

However, as shown in both Fig.\ref{fig:incompatibility} and section (a) of Fig.\ref{fig:pipeline}, we have observed that the layouts generated by VLM are commonly incompatible with the input prompt regarding object number and category.
Therefore, we introduce the second stage to refine the initial arrangement examples $\{(\hat{\poses^i}, \hat{\conds^i})\}_{i=1}^K$. 
The refinement procedure is decomposed into two critical phases through the implementation of the Chain-of-Thoughts~(CoT) strategy. 
In the first phase, we initiate the LLM~(\ie, GPT4~\cite{OpenAI2023GPT4TR}) by presenting the task introduction, user information, and the descriptions of all items on the table including the bounding box and category, provided by the detection module. This process aims to generate detailed functional requirements for the expected arrangements.
Subsequently, we prompt the LLM to infer the types and quantities of objects necessary for the functional scene, delete the redundant items, and reposition the remaining objects to generate expected coordinate-level layouts, aligning with the established functional requirements in the first phase.
To better regularize the output of the second phase, we employ an in-context learning approach which is widely used in Natural Language Processing. By presenting manually designed instances as learning material, the LLM establishes the refined layout information in a compact and uniform format, contributing to the construction of the dataset.
We defer the detailed prompt description to the project website.

% Initially, we employ the Chain-of-Thoughts~(CoT) strategy, where we inquire about the functional details within the LLM prompt, seamlessly integrating the guidelines provided by the LLM into the subsequent operational framework, thus guiding the LLM to modify the initial poses ${\hat{\poses}}$ and conditions $\hat{{\conds}}$ to align proficiently with the target poses ${\poses}$ and conditions ${\conds}$. 
% The second phase is delineated into three steps: firstly, deducing the types and quantities of objects requisite for the functional scene; secondly, removing the redundant items based on the previously gleaned information; and finally, orchestrating the remaining objects to conform to a logical and harmonious layout, in line with the predetermined specifications. 
% Furthermore, we employ an in-context learning approach: following the task description, we furnish the Large Language Model (LLM) with manually curated instances that serve as learning material. 
% These instances not only elucidate the precise format and procedure associated with our layout generation strategy but also offer explicit details regarding the positional information of each instance. This method fosters a more nuanced understanding, facilitating the model's ability to adhere to the established layout standards.

\subsection{Distilling Collected Dataset into a Diffusion Model}
\label{sec:diffusion}

We further distill the collected arrangement examples, denoted as $\data$, into a conditional generative model, $\score$. 
Our goal is to model the conditional data distribution, $\dist(\poses|\conds)$, using the conditional generative model, $\score$. 
This model is capable of generating reasonable object poses, $\poses$, that are compatible with the conditions of the objects, $\conds$. 
Consequently, we can utilize the distilled conditional generative model, $\score$, for goal specification during testing.

%%% Diffusion %%%
We employ the score-based diffusion model~\cite{SDEScoreMatching} to estimate the conditional distribution $\dist(\poses|\conds)$.
Specifically, we construct a continuous diffusion process $\{ \poses(t) \}_{t=0}^1$ indexed by a time variable $t \in [0, 1]$ using the Variance-Exploding~(VE) Stochastic Differential Equation~(SDE) proposed by~\cite{song2020score}, where $\poses(0)\sim \dist(\poses | \conds)$.
The time-indexed pose variable $\poses(t)$ is perturbed by the following SDE as $t$ increases from 0 to 1:
\begin{equation}
    d\poses = \sqrt{\frac{d[\sigma^2(t)]}{dt}} dw, \ 
    \sigma(t) = \sigma_{\text{min}}(\frac{\sigma_{\text{max}}}{\sigma_{\text{min}}})^t
\label{eq:forward_sde}
\end{equation}
where $\{w(t)\}_{t\in [0, 1]}$ is the standard Wiener process~\cite{SDEScoreMatching}, $\sigma_{\text{min}} = 0.01$ and $\sigma_{\text{max}} = 50$ are hyper-parameters. 

% Let the $p_{t}(\poses|\conds)$ denotes the marginal distribution of $\poses(t)$, 
% we aim to estimate the \textit{score function} of the perturbed conditional distribution $\nabla_{\poses} \log p_{t}(\poses|\conds)$ of all $t$, during training:
% \begin{equation}
% p_{t}(\poses(t)|\conds) = \int \mathcal{N}(\poses(t);\poses(0), \sigma^2(t)\mathbf{I}) \cdot p_0(\poses(0)|\conds) \ d\poses(0)
% \end{equation}
% It should be noted that, when $t=0$, $p_0(\poses(0)|\conds) = \dist(\poses(0)|\conds)$ is exactly the data distribution.
Let $p_{t}(\poses|\conds)$ denote the marginal distribution of $\poses(t)$. We aim to estimate the \textit{score function} of the perturbed conditional distribution $\nabla_{\poses} \log p_{t}(\poses|\conds)$ for all $t$ during training:
\begin{equation}
p_{t}(\poses(t)|\conds) = \int \mathcal{N}(\poses(t);\poses(0), \sigma^2(t)\mathbf{I}) \cdot p_0(\poses(0)|\conds) \ d\poses(0)
\end{equation}
It should be noted that when $t=0$, $p_0(\poses(0)|\conds) = \dist(\poses(0)|\conds)$, which is exactly the data distribution.

Thanks to the Denoising Score Matching~(DSM)~\cite{denosingScoreMatching}, we can obtain a guaranteed estimation of $\nabla_{\poses} p_{t}(\poses|\conds)$ by training a score network $\score: \R^{|\posespace|} \times \R^1 \times \R^{|\condspace|} \rightarrow \R^{|\posespace|}$ via the following objective $\loss(\theta)$:
\begin{equation}
\begin{aligned}
   \E_{
   t\sim \mathcal{U}(\eps, 1)}
   \left\{
   \lambda(t)\E
   \left[ \left\Vert\score(\poses(t), t | \conds)  - \frac{\poses(0) - \poses(t)}{\sigma(t)^2} \right\Vert_2^2 \right] \right\}
\end{aligned}
\label{eq:score_matching_loss}
\end{equation}
where $\poses(0) \sim \dist(\poses(0)|\conds)$ and $\poses(t) \sim \mathcal{N}(\poses(t);\poses(0), \sigma^2(t)\mathbf{I})$ $\eps$ is a hyper-parameter that denotes the minimal noise level.
When minimizes the objective in Eq.~\ref{eq:score_matching_loss}, the optimal score network satisfies $\score^*(\poses, t | \conds) = \nabla_{\poses} \log  p_{t}(\poses|\conds)$~\cite{denosingScoreMatching}.

After training, we can approximately sample goal poses from $\dist(\poses|\conds)$ by sampling from $p_{\eps}(\poses|\conds)$, as $\lim_{\eps \to 0} p_{\eps}(\poses|\conds) = \dist(\poses|\conds)$.
Sampling from $p_{\eps}(\poses|\conds)$ requires solving the following \textit{Probability Flow}~(PF) ODE~\cite{song2020score} where $\poses(1) \sim \mathcal{N}(\mathbf{0}, \sigma_{\text{max}}^2\mathbf{I})$, from $t=1$ to $t=\eps$:
\begin{equation}
\begin{aligned}
    \frac{d\poses}{dt} = - \sigma(t)\dot{\sigma}(t)\nabla_{\poses}\log p_{t}(\poses | \conds )
\label{eq:reverse_sde}
\end{aligned}
\end{equation}
where the score function $\log p_{t}(\poses | \conds )$ is empirically approximated by the trained score network $\score(\poses, t | \conds)$ and the ODE trajectory is solved by RK45 ODE solver~\cite{dormand1980family}.

%%% Implementation Details %%%
The score network $\score(\poses, t | \conds)$ is implemented as a graph neural network, so as to adapt to the varied number of input objects.
We construct all the objects as a fully-connected graph where the i-th node contains the pose $\pose_i$~(\ie, 2-D position) and the condition $\cond_i$~(\ie, 2-D sizes and 1-D label) of the i-th object.
The score network encodes the input graph by two layers of Edge-Convolution~\cite{zhang2019graph} layers and outputs the score components on each node.
The time variable $t$ is encoded into a vector by a commonly used projection layer following~\cite{song2020score}.
This vector is concatenated into intermediate features output by the Edge-Convolution layers.
We defer full training and architecture details to Appendix~\ref{appendix:architecture}.

% \yiming{Add training and architecture details here?}

\subsection{Rearrange Objects with the Trained Diffusion Model}
During test time, we first extract initial objects' conditions $\conds$ and poses $\initposes$ from the visual observation $\obs$ via GroundingDino.
We then generate goal poses $\goalposes$ for the objects with the trained score network $\score$ via Eq.~\ref{eq:reverse_sde}.

To reach the goal, we first reorder the object list by prompting GPT4 to ensure that the containers~(\eg, saucers) will be rearranged before `non-containers'~(\eg, mugs):
\begin{equation}
\begin{aligned}
\obj_j \ \text{is a container while} \ \obj_i \ \text{is not} \Rightarrow \obj_i \succ \obj_j 
\end{aligned}
\label{eq:energy_distillation_loss}
\end{equation}
Then we calculate the picking and placing points for each object using SuctionNet~\cite{cao2021suctionnet} and execute the pick-place actions. The detailed procedure is summarized in Appendix~\ref{appendix:rearrange}.

% cs_revised_v3
\begin{figure*}[t]
\begin{center}
\includegraphics[width=0.95\linewidth]{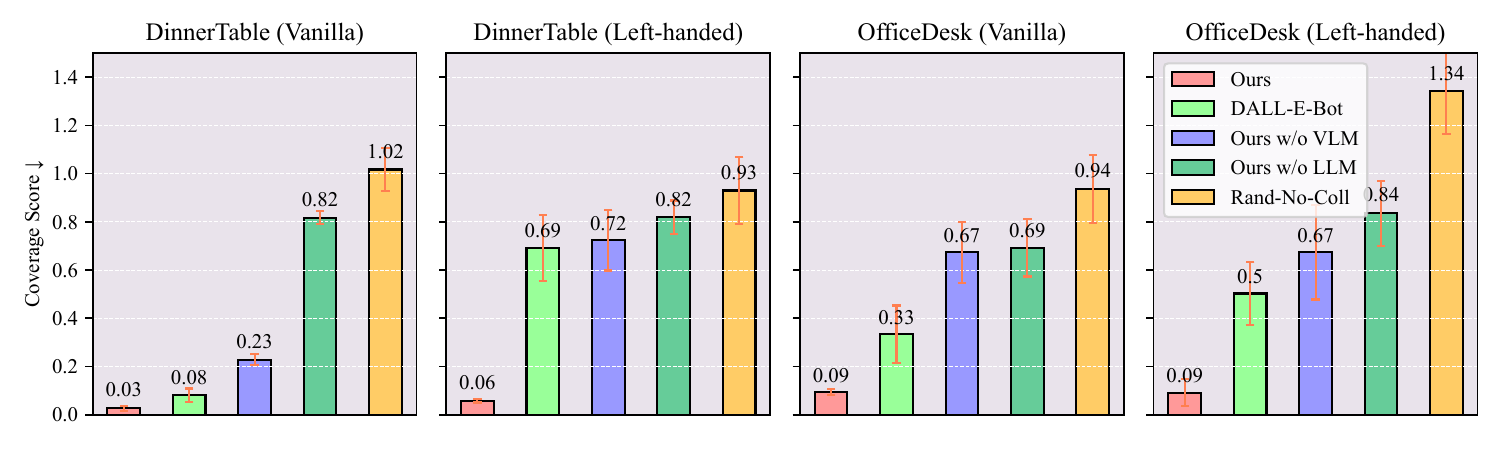}
\end{center}
\vspace{-20pt}
% \caption{Main results, dataset evaluation, coverage score, 3~4 domains(bar graphs), covering at leaset 2 scenario, 2 functional, }
\caption{Quantitative results across four domains, specifically Coverage Score bars for two functional settings (Vanilla and Left-handed) in two scenarios (Dinner table and Office desk). A lower number indicates a smaller deviation between the generated layouts and the ground truth layouts, signifying better performance. The mean and standard deviation are reported for comparison among our methods and baseline algorithms.}
\label{fig:coverage_score}
\vspace{-10pt}
\end{figure*}

% \vspace{-10pt}
\section{EXPERIMENTS}
Similar to~\cite{dallebot}, we evaluate our method using both subjective and objective metrics. In Sec~\ref{sec:setup}, we will first introduce the domains and baselines used in our experiments. In Sec~\ref{sec:objective}, we collect a test set that contains 30 ground truth arrangements for each domain and evaluate the effectiveness of each method in generating proper arrangements that satisfy different functional requirements under various configurations. In Sec~\ref{sec:subjective}, we conduct real-world rearrangement experiments at scale and measure the quality of the rearrangement results through user ratings.
We defer the detailed experimental setups and implementation details to supplementary.

\subsection{Setups}
\label{sec:setup}
\textbf{Domains.}
% We consider four domains formed by the combination of two scenarios, \ie, dinner table and office desk, and two functionalities, \ie, ``setting table for right-handed people'' (Vanilla) and ``setting table for left-handed people'' (Left-handed): \textit{DinnerTable (Vanilla)}, \textit{DinnerTable (Left-handed)}, \textit{OfficeDesk (Vanilla)} and \textit{OfficeDesk (Left-handed)}, where we generate arrangement 969, 132, 165 and 161 examples via our autonomous data generation pipeline for training the diffusion model, respectively.
We consider four domains formed by the combination of two scenarios, namely, the dinner table and office desk, and two functionalities, which are ``setting the table for right-handed people'' (Vanilla) and ``setting the table for left-handed people'' (Left-handed). These domains are labeled as follows: \textit{DinnerTable (Vanilla)}, \textit{DinnerTable (Left-handed)}, \textit{OfficeDesk (Vanilla)}, and \textit{OfficeDesk (Left-handed)}. We generated 969, 132, 165, and 161 examples, respectively, for the four domains mentioned above using our autonomous data generation pipeline to train the diffusion model.

\textbf{Baselines.} 
% We compare our method with a zero-shot baseline, \textit{DALL-E-Bot}, that integrates a VLM into the rearrangement pipeline such that do not depend on a dataset that requires human annotation or designation.
% Following~\cite{dallebot}, we also compare our method with \textit{Rand-No-Coll} baseline that places objects randomly in the environment while ensuring they do not overlap.
We compare our method with \textit{DALL-E-Bot}, a zero-shot baseline that incorporates a VLM and a rejection sampling strategy, which repeatedly samples the goal layout until it successfully passes the filters \textit{(1. No duplicate items of the same category. 2. No overlap between objects)} into the rearrangement pipeline, eliminating the need for a dataset requiring human annotation or designation. Additionally, following the approach in~\cite{dallebot}, we evaluate our method against the \textit{Rand-No-Coll} baseline, which randomly places objects in the environment while ensuring they do not overlap.

% \textbf{Metrics.} For comprehensive evaluation, we introduce two typical metrics, one subjective and one objective, to evaluate the rearrangement results. \textit{Coverage Score} measures the diversity and fidelity of the rearrangement results by calculating the Minimal-Matching-Distance~(MMD)~\cite{MMD} between $S_{T}$ and a fixed set of ground-truth examples $S_{gt}$ from $\tar$: $\sum\limits_{\s_{gt} \in S_{gt}} \min\limits_{\s_T \in S_{T}} ||\s_{gt} - \s_T||$.
% \textit{User-rating} 

\subsection{Arrangement Evaluation}
\label{sec:objective}

\begin{figure}[t]
\begin{center}
\includegraphics[width=0.95\linewidth]{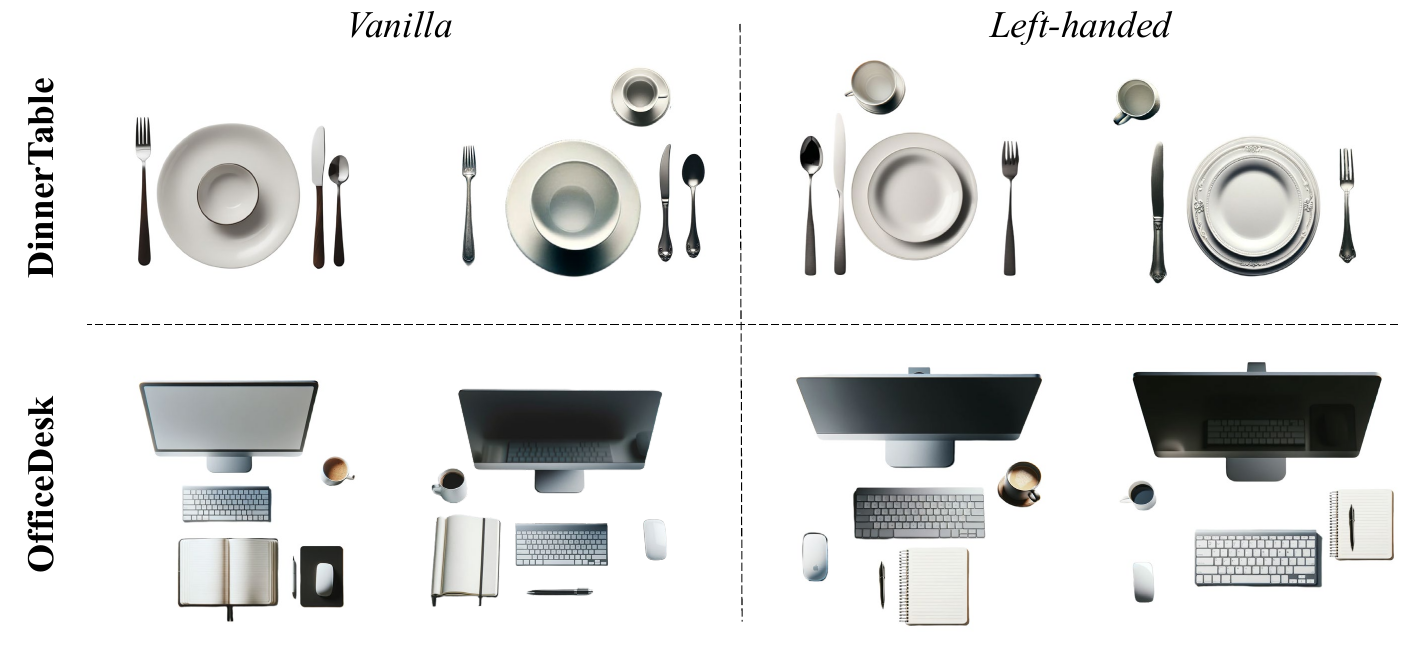}
\end{center}
\vspace{-15pt}
\caption{Visualization of test set examples: We randomly pick 2 test examples from each domain. To visualize test examples (\ie, LLM-refined bounding boxes), we crop the mask of each object from the results generated by VLM, and then move the object mask into the LLM refined bounding box.}
\label{fig:examples}
\vspace{-15pt}
\end{figure}

\begin{figure*}[tbp]
\begin{center}
\includegraphics[width=\linewidth]{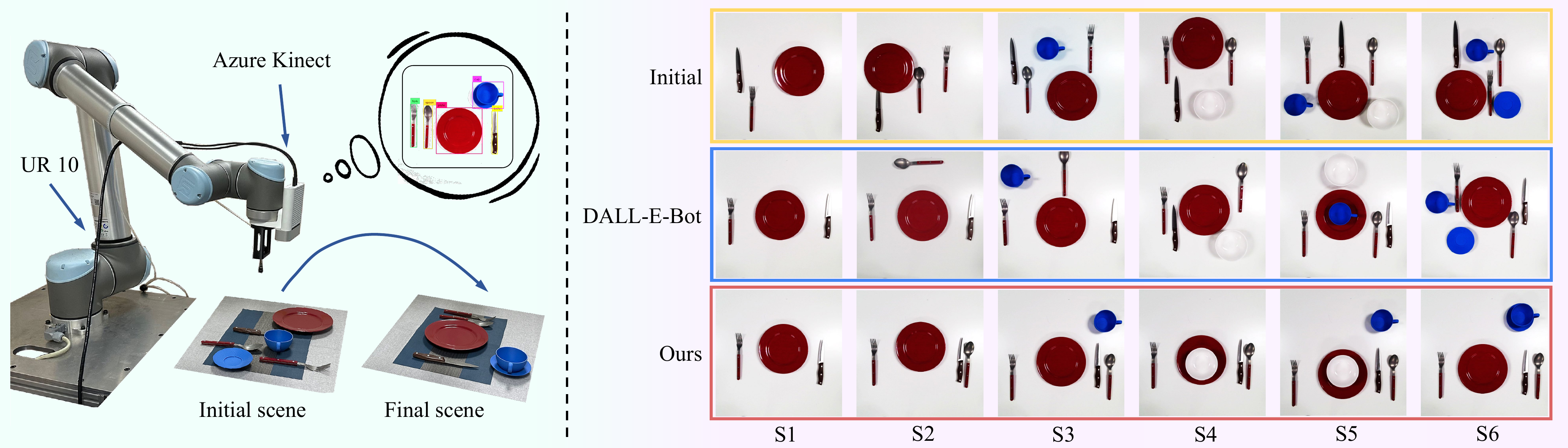}  % wmd version
\end{center}
\label{-20pt}
\caption{\textbf{Left:} The real-world setup for object rearrangement. \textbf{Right:} Qualitative results of real-world experiments on \textit{DinnerTable~(Vanilla)} scenario. We design 6 different scenes~(\ie, $S1\sim S6$) with increasing complexity.}
\label{fig:realworld_qualitative}
\vspace{-15pt}
\end{figure*}

% In the following experiments, we focus on evaluating the effectiveness of generating proper arrangement goals and do not consider the rearrangement process. For each domain, we generate arrangements for the 30 configurations from the test set $\poses_{gt}$, which consists of user-preferred layouts, manually designed to be well-organized and meet the functional requirements of various scenarios, none of which is included in the training set. Examples of both training and test datasets are provided in Fig~\ref{fig:examples}.
In the following experiments, we focus on evaluating the effectiveness of generating proper arrangement goals and do not consider the rearrangement process. For each domain, we generate arrangements for the 30 configurations from the test set $\poses_{gt}$, which consists of user-preferred layouts that are well-organized and meet the functional requirements of various scenarios, none of which are included in the training set. Examples from the test dataset are provided in Fig~\ref{fig:examples}.

In the case of \textit{DALL-E-Bot}, we take the target poses after the ICP-matching as its arrangement results since we do not provide an initial state. 
To evaluate the generated arrangement goals $\poses_{gen}$, we employ the \textit{Coverage Score} proposed by~\cite{wu2022targf}, which measures the diversity and fidelity of the rearrangement results by calculating the Minimal-Matching-Distance~(MMD)~\cite{MMD} between $\poses_{gen}$ and $\poses_{gt}$: 
\begin{equation}
    \sum\limits_{\pose_{gt} \in \poses_{gt}} \min\limits_{\pose_{gen} \in \poses_{gen}} ||\pose_{gt} - \pose_{gen}||^2_2.
\end{equation}
% \mingdong{1.metrics clarification, a larger/smaller number indicates XXXX}

% As shown in Fig.~\ref{fig:coverage_score}, our method significantly outperforms all the baselines across four domains, which showcases the effectiveness of our method to generate arrangement goals.
% The lower-bound method \textit{Rand-No-Coll} achieves the worst performance consistently, which showcases the effectiveness of the \textit{Coverage Score} metric.
% Notably, in the \textit{Left-handed} domains, our method exhibits a greater advantage compared to DALL-E-Bot than \textit{Vanilla}. This is because VLM struggles to align with the functional requirements in the prompt, whereas our approach, which refines VLM using LLM, ensures that the training data better aligns with the functional requirements.

As shown in Fig.~\ref{fig:coverage_score}, our method significantly outperforms all the baselines across four domains, showcasing the effectiveness of our method in generating arrangement goals. 
The lower-bound method, \textit{Rand-No-Coll}, consistently achieves the worst performance, highlighting the effectiveness of the \textit{Coverage Score} metric.
Notably, in all the \textit{Left-handed} domains, our method exhibits a greater advantage compared to DALL-E-Bot than \textit{Vanilla}'s. This is because VLM struggles to align with the functional requirements in the prompt, whereas our approach, which refines VLM using LLM, ensures that the training data better aligns with the functional requirements.

\begin{table}[h]
\centering
\caption{Analyze the marginal KL-divergence on \textit{DinnerTable~(Left-handed)}.}
\vspace{-5pt}
\resizebox{\linewidth}{!}{
\input{tabs/marginal_kld}
}
% \vspace{-15pt}
\label{table:kld}
\end{table}

% To get an in-depth understanding of our advantages, we further conduct marginal statistical analysis on \textit{DinnerTable (Left-handed)}.
% Specifically, we propose \textit{marginal-KL-analysis}, which calculates the relative positional distribution of two specific categories~(\eg, Plate2Fork) of items within the results generated by various methods and determines the KL-distance between this distribution and the one observed in the test dataset.
% We defer the metric details into Appendix~\ref{appendix:marginal_kld}.

To gain an in-depth understanding of our advantages, we conduct further marginal statistical analysis on \textit{DinnerTable (Left-handed)}. Specifically, we introduce \textit{marginal-KL-analysis}, which calculates the relative positional distribution of two specific categories (\eg, Plate2Fork) of items within the results generated by various methods and determines the KL-distance between this distribution and the one observed in the test dataset. Detailed information about this metric is deferred to Appendix~\ref{appendix:marginal_kld}.

As shown in Table~\ref{table:kld}, our method outperforms the baselines in most cases. In \textit{Plate2Spoon}, our method still achieves comparable performance to the \textit{DALL-E-Bot}.
% \mingdong{Enrich KL-Div details here}
% These indicate that 

Notably, both VLM and LLM play crucial roles in our method's performance. 
As demonstrated in Fig.\ref{fig:coverage_score} and Table\ref{table:kld}, our method experiences a significant performance drop when either LLM or VLM is ablated, to the extent that it falls significantly below DALL-E-Bot.
% Similarly, in Table~\ref{table:kld}, 

% \begin{wraptable}{tr}{0.40\textwidth}
%     \centering
%     % \vspace{-10pt}
%     \input{tabs/inference_time}
%     \caption{Inference Time}
%     % \vspace{-10pt}
%     \label{table:inference_time}
%     % \vspace{-5pt}
% \end{wraptable}

% \begin{figure}[tbp]
% \begin{center}
% \includegraphics[width=\linewidth]{figs/real_world.pdf}
% \end{center}
% \caption{\mingdong{Shall we need?} Real-world setup. Similar to the dallebot. Left: initial, Right: Final}
% \label{fig:realworld_setup}
% \end{figure}

\subsection{Real World Rearrangement Experiments}
\label{sec:subjective}

In the following experiments, we focus on evaluating the effectiveness of our method in real-world rearrangement. 
We compare our method with the most competitive baseline, \textit{DALL-E-Bot}, as verified in objective evaluations (see Section~\ref{sec:objective}). 
To ensure a fair comparison, we designed 6 different scenes~(denoted as $S1\sim S6$, the scene complexity increase from $S1$ to $S6$) with 3 different initializations within \textit{DinnerTable (Vanilla)} domain and deployed both methods on the same set of initial configurations.
As depicted in Fig.\ref{fig:realworld_qualitative} (a), the real-world experiments were conducted using a UR10 robot arm equipped with an Azure Kinect RGBD camera. 
Following the evaluation setup used in many other studies on arrangement generation~\cite{neatnet, dallebot}, we conducted a user study to evaluate the rearrangement results, as there are no ground truth arrangements in the real-world setting.
The user is asked to score from 1~(very bad) to 10~(very good), according to their preferences and functional requirements.
We recruited 30 users, including both male and female, with ages ranging from 18 to 55. Each rearrangement result is assessed by all the users, resulting in a total of 900 ratings.

As depicted in Fig.~\ref{fig:realworld_qualitative} (b), \textit{DALL-E-Bot} performs well in the scenes with low complexity (\eg, S1 and S2). However,  as the complexity of the scene increases, \textit{DALL-E-Bot} experiences a decline in performance, since VLM-generated images could not pass the filer, the object matching module and ICP are always affected by distracting objects, whereas our approach maintains stable performance and remarkably handles the S6 scene, which includes duplicate plates of different sizes and a stacked plate-cup set.
This indicates that our distilled rearrangement priors can adapt to varying object numbers and categories, thanks to the graph neural network design, and can effectively rearrange objects into plausible results.

\begin{figure}[t]
\begin{center}
\includegraphics[width=\linewidth]{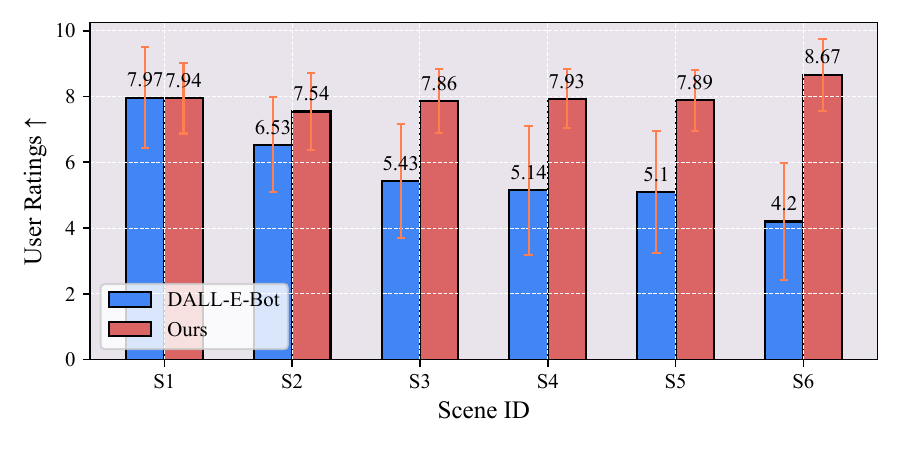}
\end{center}
\vspace{-15pt}
\caption{User ratings for each method on real-world arrangements. Each bar demonstrates the mean and standard deviation across all users.}
\vspace{-15pt}
\label{fig:realworld_quantitative}
\end{figure}

The results presented in Fig.~\ref{fig:realworld_quantitative} consistently demonstrate this phenomenon: the average user rating for \textit{DALL-E-Bot} decreases from S1 to S6, whereas our method not only maintains its performance but also exhibits slight improvement.
These findings highlight the effectiveness of our approach in generating compatible and robust goals for object rearrangement.

% \begin{wraptable}{tr}{0.50\linewidth}
%     \centering
%     % \vspace{-10pt}
%     \tiny
%     \input{tabs/inference_time}
%     \caption{Inference Time}
%     % \vspace{-10pt}
%     \label{tab:inference_time}
%     % \vspace{-5pt}
% \end{wraptable}
% Finally, our approach boasts greater time efficiency, owing to our utilization of distillation techniques that enable the acquisition of more streamlined and effective representations.

% DALL-E-Bot needs a filter to remove failed images generated by VLM.
% \begin{table}[t]
% \vspace{-5pt}
% \centering
% \caption{Model parameters}
% \vspace{-5pt}
% \input{tabs/inference_time}
% \vspace{-10pt}
% \label{table:inference_time}
% \end{table}

\begin{figure}[t]
\begin{center}
\includegraphics[width=\linewidth]{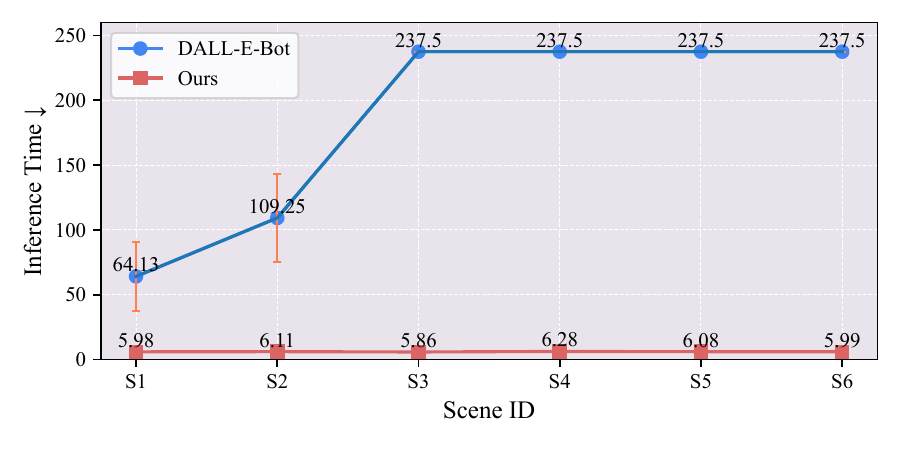}
\end{center}
\vspace{-15pt}
\caption{\textbf{Inference time} for each method on real-world arrangements. The complexity of the scene gradually increases from S1 to S6.}
\vspace{-15pt}
\label{fig:inference time}
\end{figure}

Finally, our approach boasts greater time efficiency because we distill knowledge from the large model~(3.5B parameters for StableDiffusion XL) into a lightweight compact representation~(180K parameters for ours), avoiding repeated inference of the large model at test time. 
To illustrate this point, we evaluated of the average inference times required for goal specification in our method as compared to \textit{DALL-E-Bot} on the \textit{DinnerTable~(Vanilla)} dataset, using 10 identical initializations.
As detailed in Fig. \ref{fig:inference time}, our method achieves a substantial advantage in terms of efficiency. This advantage arises because VLM-generated images often struggle to pass the filters designed by \textit{DALL-E-Bot} when dealing with complex scenes. In the context of S3-S6, marked by higher complexity, \textit{DALL-E-Bot} consistently fails to pass the filter and exceeds the designated filtration budget ($\leq 10$ sampling times), resulting in the same long inference time. In contrast, our method consistently generates compatible layouts in a few seconds.

% \vspace{-5pt}
\section{CONCLUSION}
% \vspace{-5pt}

We are the first to distill functional rearrangement priors from large models into compact representations for object rearrangement tasks. Specifically, we propose a novel approach that collects diverse arrangement examples using both LLMs and VLMs and then distills the examples into a diffusion model. This approach leverages the scalability and generalization of large models to facilitate the distilled diffusion model in generating goals compatible with the initial configuration, effectively addressing compatibility issues. 
% \mingdong{tone down the language about the results}
Our results, including real-world experiments, go beyond the performance of the baselines in more complex scenarios. Furthermore, our further analysis suggests that both VLM and LLM play crucial roles in the performance of our method.

Existing limitations of this work include: 
1) Lack of rotation: In this work, we focus on validating the effectiveness of distilling a 2-D layout dataset for rearrangement and only consider 2-D translation for simplicity. Future work could involve distilling unambiguous orientation knowledge from large models once a well-developed semantic orientation-aware detection model is available for everyday objects.
2) One model per function: Our approach requires training a specific model for each corresponding functional requirement. 
Future work could involve extending this approach to a more generalized model capable of handling diverse daily demands. 

% 2) One model one scenario: our methods Future work could involve integrating a rotation-aware perception module and incorporating language conditions to further enhance the performance and capabilities of our method.}

% {
% \bibliographystyle{IEEEtran}
% \bibliography{IEEEabrv,reference}
% }

% \subsection{Datasets}

% \begin{figure}[t]
%   \centering
% \includegraphics[trim=60 60 300 15,clip, scale=0.64]{figures/object.pdf}
%   \caption{\textbf{Object Examples from PartNet-Mobility Dataset.}}
%   \label{fig:dataset}
% \end{figure}

% For \textit{pull drawer} and \textit{push drawer} tasks, the objects are chosen from 4 subcategoreis categories, we randomly select 10-12 objects in each category for training (44 objects in total) and 3 objects for testing (12 objects in total). 

% For \textit{open door} and \textit{close door} tasks, the objects are chosen from 2 subcategoreis categories, we randomly select 22 objects in each category for training (44 objects in total) and 8 objects for testing (16 objects in total). 

% For \textit{pick and place}, the ojects are chosen from 6 subcategoreis, we randomly select 8 objects in each category for training and 2 for testing. There are 4 tables for placement. The daily items on the table, including Pot, Bread Machine and Bottle, are randomly placed. Considering all the combinations of table and daily object placements, we use 48 different arrangements.

% \clearpage

\section{APPENDIX}

\subsection{Details of Marginal KL-divergence Analysis}
\label{appendix:marginal_kld}

Marginal KL-Divergence is employed to assess the relative 2D positional distribution of two specific categories. Initially, we performed the following 2D Gaussian Kernel Density Estimation (KDE) function to estimate the probability density of 2D positional distances between the two categories:
\begin{equation}
    \hat{f}(x, y) = \frac{1}{n h_x h_y} \sum_{i=1}^{n} K\left(\frac{x - x_i}{h_x}\right) \cdot K\left(\frac{y - y_i}{h_y}\right),
\label{eq:kde}
\end{equation}

where $(x, y)$ is the 2D relational distance of two points, $n$ is the number of samples, $h_x$ and $h_y$ are bandwidth parameters in the x and y directions, respectively, and $K(u) = \frac{1}{2\pi} e^{ - \frac{u^2}{2}}$ is the kernel function, chosen as the probability density function of the standard normal distribution.
Subsequently, we employed the following Kullback-Leibler Divergence (KLD) to measure the difference between the two probability distributions:

\begin{equation}
    D_{KL}(P \, \| \, Q) = \sum_{i} P(i) \cdot \log\left(\frac{P(i)}{Q(i)}\right),
\end{equation}

where event $i$ represents data points of 2D positional distances between two specific categories, $P(i)$ and $Q(i)$ denote the probabilities of event $i$, computed from Eq. \ref{eq:kde}.

% \subsection{Prompt Engineering for Data Collection}
% \label{appendix:data}

\vspace{-5pt}
\subsection{Training and Architectural Details of the Score Network}
\label{appendix:architecture}

%%%%%%%%%%%%%%%%%%%%%%%%%% Network Structure %%%%%%%%%%%%%%%%%%%%%%%%%%%%%%%%%%
\begin{figure}[h]
\begin{center}
\includegraphics[width=\linewidth]{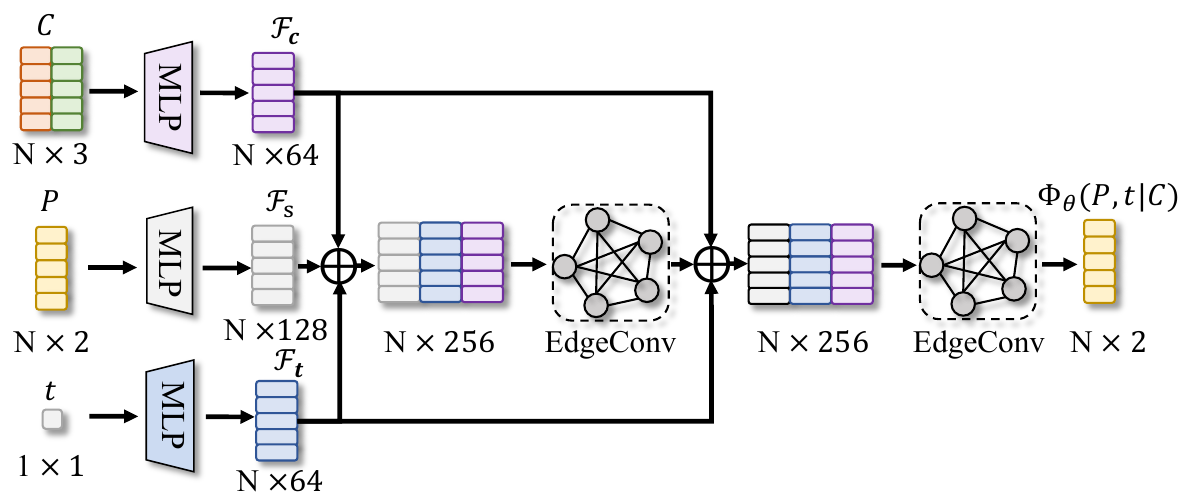}
\end{center}
% \vspace{-10pt}
\vspace{-5pt}
\caption{\textbf{Architecture of the Score Network}}
\vspace{-5pt}
\label{fig:network structure}
% \vspace{-15pt}
\end{figure}
%%%%%%%%%%%%%%%%%%%%%%%%% Network Structure %%%%%%%%%%%%%%%%%%%%%%%%%%%%%%%%%%%%%%%%%%%%

\textbf{Training Details:} We use the Adam optimizer with a learning rate of 2e-4 for training, and the batch size is set to 16. It takes 9 hours to train on a single RTX 3090 for primitive policy to converge. The number of steps $\epsilon$ is configured to 500 in the sampling process.

\vspace{-5pt}
\subsection{Test Time Rearrangement Algorithm}
\vspace{-5pt}
\label{appendix:rearrange}
\begin{algorithm}
\caption{Test Time Rearrangement Algorithm}
\label{alg:rearrange}
\begin{algorithmic}[1]
    \State \textbf{Initialization:} Learned score network $\score$, number of objects $N$, received visual observation $\obs$
    \State Extract initial object poses $\initposes$ and conditions $\conds$ from $\obs$
    \State Generate goal poses $\goalposes \sim \dist(\poses | \conds)$ using Eq.~\ref{eq:reverse_sde}
    \State Reorder the object list $\{o_i\}$ according to 
    \For{$i=1$ {\bfseries to} $N$}
    \State Compute translation $\trans_i$ for $\obj_i$ using $\initpose_i$ and $\goalpose_i$
    \For{$j=i+1$ {\bfseries to} $N$}
    \If{$o_i$ will collide with $o_j$ after translation}
        \State Move $o_j$ away
    \EndIf
    \EndFor
    \State Execute pick-and-place actions via SuctionNet
    % \State Get the picking point $\pose_{\text{pick}}$ by SuctionNet
    % \State Get the placing point $\pose_{\text{place}}$ according to $\pose_{\text{pick}}$ and $\trans_i$
    % \State Execute picking $\pose_{\text{pick}}$ and placing $\pose_{\text{place}}$
    \EndFor
\end{algorithmic}
\end{algorithm}
\label{sec:rearrange}

% \subsection{Examples of Training and Test Dataset}
% \label{appendix:DataExample}

% \subsection{Dall-E-Bot Implementation}
% \label{appendix:baselines}
% \clearpage

{
\bibliographystyle{IEEEtran}
\bibliography{IEEEabrv,reference}
}

\end{document}

%% file: notations.tex
% Domain relevant
\newcommand{\sset}{\mathcal{S}}
\newcommand{\frees}{\mathcal{S}_f}
\newcommand{\aset}{\mathcal{A}}
\newcommand{\eps}{\epsilon}
\newcommand{\init}{\rho_0}
\newcommand{\E}{\mathbb{E}}
\newcommand{\R}{\mathbb{R}}
\newcommand{\hS}{\mathbb{S}}
\newcommand{\hP}{\mathbb{P}}
\newcommand{\M}{\mathcal{M}}
\newcommand{\bs}{\vb*{bs}}
\newcommand{\N}{\mathcal{N}}
\newcommand{\G}{\mathcal{G}}
\newcommand{\C}{\mathcal{C}}
\newcommand{\Env}{\mathcal{E}}
\newcommand{\x}{\vb*{x}}

\newcommand{\obj}{o}
\newcommand{\objs}{O}
\newcommand{\pose}{p}
\newcommand{\conds}{C}
\newcommand{\poses}{P}
\newcommand{\initpose}{\pose^0}
\newcommand{\initposes}{\poses^0}
\newcommand{\goalposes}{\poses^g}
\newcommand{\goalpose}{\pose^g}
\newcommand{\cond}{c}
\newcommand{\size}{s}
\newcommand{\cate}{y}
\newcommand{\mask}{m}
\newcommand{\obs}{I_{\objs}}
\newcommand{\func}{f}
\newcommand{\dist}{p_{\func}}
\newcommand{\data}{\textit{D}_{\func}}
\newcommand{\score}{\vb*{\Phi}_{\theta}}
\newcommand{\posespace}{\mathcal{P}}
\newcommand{\condspace}{\mathcal{C}}

\newcommand\mydata[2]{$#1_{\pm#2}$}
\newcommand{\indicator}{\mathbbm{1}}
\newcommand{\loss}{\mathcal{L}}
\newcommand{\trans}{\mathcal{T}}

% \newcommand{\trans}{T}
% \newcommand{\rot}{R}
% \newcommand{\cand}{\hat{\pose}}
% \newcommand{\quat}{\vb*{q}}
% % \newcommand{\data}{\mathcal{D}}
% % \newcommand{\dist}{p_{\text{data}}}
% \newcommand{\energy}{\vb*{\Psi}_{\phi}}
% \newcommand{\score}{\vb*{\Phi}_{\theta}}
% \newcommand{\energyscore}{\vb*{\Phi}_{\phi}}
% \newcommand{\percent}{\delta}
% \newcommand{\Z}{\vb*{z}}
% \newcommand{\posespace}{\mathcal{P}}

% \newcommand{\indicator}{\mathbbm{1}}
% \newcommand{\joint}{\vb*{J}}
% \newcommand{\base}{\vb*{b}}
% \newcommand{\pbase}{\base_{p}}
% \newcommand{\qbase}{\base_{q}}
% \newcommand{\hist}{H}
% \newcommand{\unitvec}{\vec{\vb*{1}}}
% \newcommand{\ppolicy}{\pi_p^{\theta}}
% \newcommand{\rpolicy}{\pi_r^{\phi}}
% \newcommand{\policy}{\pi^{\theta, \phi}}
% \newcommand{\augmentation}{\mathscr{A}}

% \newcommand{\z}{\vb*{z}}
% \newcommand{\ac}{\vb*{a}}
% \newcommand{\s}{\vb*{s}}
% \newcommand{\vel}{\vb*{v}}
% \newcommand{\g}{\vb*{g}}
% \newcommand{\fprox}{\vb*{F}_{proxy}}
% \newcommand{\col}{\vb*{c}}
% \newcommand{\gtar}{\g_{tar}}
% \newcommand{\gsup}{\g_{sup}}
% \newcommand{\gdual}{\g_{dual}}
% \newcommand{\supp}{p_{sup}}
% \newcommand{\tarD}{S^*_{tar}}
% \newcommand{\supD}{S^*_{sup}}
% \newcommand{\phiS}{\vb*{\Phi}_{\theta}}
% \newcommand{\tarS}{\vb*{\Phi}_{tar}^{\theta}}
% \newcommand{\supS}{\vb*{\Phi}_{sup}^{\phi}}
% \newcommand{\tarE}{\vb*{\Psi}_{tar}^{\theta}}
% \newcommand{\supE}{\vb*{\Psi}_{sup}^{\phi}}
% \newcommand{\dualS}{\vb*{\Phi}_{dual}}
% \newcommand{\bt}{\vb*{t}}

\def\eg{\emph{e.g}.} \def\Eg{\emph{E.g}.}
\def\ie{\emph{i.e}.} \def\Ie{\emph{I.e}.}
\def\cf{\emph{c.f}.} \def\Cf{\emph{C.f}.}
\def\etc{\emph{etc}.} \def\vs{\emph{vs}.}
\def\wrt{w.r.t. } \def\dof{d.o.f. }
\def\etal{\emph{et al}. }

%% file: tabs/marginal_kld.tex
% \begin{tabular}{lccccc}
% \hline
%             & Ours  & DALL-E-Bot & Rand-No-Coll & Ours w/o LLM & Ours w/o VLM \\ \hline
% Plate2Fork  & \textbf{6.72}  & 73.48    & 74.45    & 72.29        & 78.46        \\ \hline
% Plate2Knife & \textbf{6.81}  & 32.41    & 34.21    & 35.17        & 27.42        \\ \hline
% Plate2Spoon & 12.59 & \textbf{11.52}    & 12.40    & 6.28         & 8.94         \\ \hline
% \end{tabular}

\begin{tabular}{lccc}
\toprule
             & Plate2Fork$ \ \downarrow$    & Plate2Knife$\ \downarrow$   & Plate2Spoon$\ \downarrow$    \\
\midrule
Rand-No-Coll & 74.45         & 34.21         & 12.40          \\
DALL-E-Bot   & 73.48         & 32.41         & \textbf{11.52} \\
% \midrule
Ours         & \textbf{6.72} & \textbf{6.81} & 12.59          \\
\midrule
Ours w/o LLM & 72.29         & 35.17         & 6.28           \\
Ours w/o VLM & 78.46         & 27.42         & 8.94           \\
\bottomrule
\end{tabular}

%% file: main_RAL.bbl
\begin{thebibliography}{10}
\providecommand{\url}[1]{#1}
\csname url@samestyle\endcsname
\providecommand{\newblock}{\relax}
\providecommand{\bibinfo}[2]{#2}
\providecommand{\BIBentrySTDinterwordspacing}{\spaceskip=0pt\relax}
\providecommand{\BIBentryALTinterwordstretchfactor}{4}
\providecommand{\BIBentryALTinterwordspacing}{\spaceskip=\fontdimen2\font plus
\BIBentryALTinterwordstretchfactor\fontdimen3\font minus \fontdimen4\font\relax}
\providecommand{\BIBforeignlanguage}[2]{{%
\expandafter\ifx\csname l@#1\endcsname\relax
\typeout{** WARNING: IEEEtran.bst: No hyphenation pattern has been}%
\typeout{** loaded for the language `#1'. Using the pattern for}%
\typeout{** the default language instead.}%
\else
\language=\csname l@#1\endcsname
\fi
#2}}
\providecommand{\BIBdecl}{\relax}
\BIBdecl

\bibitem{rearrange}
D.~Batra, A.~X. Chang, S.~Chernova, A.~J. Davison, J.~Deng, V.~Koltun, S.~Levine, J.~Malik, I.~Mordatch, R.~Mottaghi \emph{et~al.}, ``Rearrangement: A challenge for embodied ai,'' \emph{arXiv preprint arXiv:2011.01975}, 2020.

\bibitem{wu2022targf}
M.~Wu, F.~Zhong, Y.~Xia, and H.~Dong, ``{TarGF}: Learning target gradient field for object rearrangement,'' \emph{arXiv preprint arXiv:2209.00853}, 2022.

\bibitem{shen2020structformer}
Y.~Shen, Y.~Tay, C.~Zheng, D.~Bahri, D.~Metzler, and A.~Courville, ``Structformer: Joint unsupervised induction of dependency and constituency structure from masked language modeling,'' \emph{arXiv preprint arXiv:2012.00857}, 2020.

\bibitem{liu2022structdiffusion}
W.~Liu, T.~Hermans, S.~Chernova, and C.~Paxton, ``Structdiffusion: Object-centric diffusion for semantic rearrangement of novel objects,'' \emph{arXiv preprint arXiv:2211.04604}, 2022.

\bibitem{simeonov2023shelving}
A.~Simeonov, A.~Goyal, L.~Manuelli, L.~Yen-Chen, A.~Sarmiento, A.~Rodriguez, P.~Agrawal, and D.~Fox, ``Shelving, stacking, hanging: Relational pose diffusion for multi-modal rearrangement,'' \emph{arXiv preprint arXiv:2307.04751}, 2023.

\bibitem{neatnet}
I.~Kapelyukh and E.~Johns, ``My house, my rules: Learning tidying preferences with graph neural networks,'' in \emph{5th Annual Conference on Robot Learning}, 2021.

\bibitem{dallebot}
I.~Kapelyukh, V.~Vosylius, and E.~Johns, ``Dall-e-bot: Introducing web-scale diffusion models to robotics,'' \emph{IEEE Robotics and Automation Letters}, 2023.

\bibitem{rombach2022high}
R.~Rombach, A.~Blattmann, D.~Lorenz, P.~Esser, and B.~Ommer, ``High-resolution image synthesis with latent diffusion models,'' in \emph{Proceedings of the IEEE/CVF conference on computer vision and pattern recognition}, 2022, pp. 10\,684--10\,695.

\bibitem{SDEScoreMatching}
\BIBentryALTinterwordspacing
Y.~Song, J.~Sohl-Dickstein, D.~P. Kingma, A.~Kumar, S.~Ermon, and B.~Poole, ``Score-based generative modeling through stochastic differential equations,'' in \emph{International Conference on Learning Representations}, 2021. [Online]. Available: \url{https://openreview.net/forum?id=PxTIG12RRHS}
\BIBentrySTDinterwordspacing

\bibitem{lee2023aligning}
K.~Lee, H.~Liu, M.~Ryu, O.~Watkins, Y.~Du, C.~Boutilier, P.~Abbeel, M.~Ghavamzadeh, and S.~S. Gu, ``Aligning text-to-image models using human feedback,'' \emph{arXiv preprint arXiv:2302.12192}, 2023.

\bibitem{black2023training}
K.~Black, M.~Janner, Y.~Du, I.~Kostrikov, and S.~Levine, ``Training diffusion models with reinforcement learning,'' \emph{arXiv preprint arXiv:2305.13301}, 2023.

\bibitem{liu2023grounding}
S.~Liu, Z.~Zeng, T.~Ren, F.~Li, H.~Zhang, J.~Yang, C.~Li, J.~Yang, H.~Su, J.~Zhu \emph{et~al.}, ``Grounding dino: Marrying dino with grounded pre-training for open-set object detection,'' \emph{arXiv preprint arXiv:2303.05499}, 2023.

\bibitem{wei2022chain}
J.~Wei, X.~Wang, D.~Schuurmans, M.~Bosma, F.~Xia, E.~Chi, Q.~V. Le, D.~Zhou \emph{et~al.}, ``Chain-of-thought prompting elicits reasoning in large language models,'' \emph{Advances in Neural Information Processing Systems}, vol.~35, pp. 24\,824--24\,837, 2022.

\bibitem{abdo2015robot}
N.~Abdo, C.~Stachniss, L.~Spinello, and W.~Burgard, ``Robot, organize my shelves! tidying up objects by predicting user preferences,'' in \emph{2015 IEEE International Conference on Robotics and Automation (ICRA)}.\hskip 1em plus 0.5em minus 0.4em\relax IEEE, 2015, pp. 1557--1564.

\bibitem{schuster2012learning}
M.~J. Schuster, D.~Jain, M.~Tenorth, and M.~Beetz, ``Learning organizational principles in human environments,'' in \emph{2012 IEEE International Conference on Robotics and Automation}.\hskip 1em plus 0.5em minus 0.4em\relax IEEE, 2012, pp. 3867--3874.

\bibitem{abdo2016organizing}
N.~Abdo, C.~Stachniss, L.~Spinello, and W.~Burgard, ``Organizing objects by predicting user preferences through collaborative filtering,'' \emph{The International Journal of Robotics Research}, vol.~35, no.~13, pp. 1587--1608, 2016.

\bibitem{fisher2012example}
M.~Fisher, D.~Ritchie, M.~Savva, T.~Funkhouser, and P.~Hanrahan, ``Example-based synthesis of 3d object arrangements,'' \emph{ACM Transactions on Graphics (TOG)}, vol.~31, no.~6, pp. 1--11, 2012.

\bibitem{MakeItHome2011}
L.~F. Yu, S.~K. Yeung, C.~K. Tang, D.~Terzopoulos, T.~F. Chan, and S.~J. Osher, ``Make it home: automatic optimization of furniture arrangement,'' \emph{ACM Transactions on Graphics (TOG)-Proceedings of ACM SIGGRAPH 2011, v. 30,(4), July 2011, article no. 86}, vol.~30, no.~4, 2011.

\bibitem{RJMCMC2012}
Y.-T. Yeh, L.~Yang, M.~Watson, N.~D. Goodman, and P.~Hanrahan, ``Synthesizing open worlds with constraints using locally annealed reversible jump mcmc,'' \emph{ACM Transactions on Graphics (TOG)}, vol.~31, no.~4, pp. 1--11, 2012.

\bibitem{SceneNet2016}
A.~Handa, V.~P{\u{a}}tr{\u{a}}ucean, S.~Stent, and R.~Cipolla, ``Scenenet: An annotated model generator for indoor scene understanding,'' in \emph{2016 IEEE International Conference on Robotics and Automation (ICRA)}.\hskip 1em plus 0.5em minus 0.4em\relax IEEE, 2016, pp. 5737--5743.

\bibitem{songchun2018}
S.~Qi, Y.~Zhu, S.~Huang, C.~Jiang, and S.-C. Zhu, ``Human-centric indoor scene synthesis using stochastic grammar,'' in \emph{Proceedings of the IEEE Conference on Computer Vision and Pattern Recognition}, 2018, pp. 5899--5908.

\bibitem{wei2023lego}
Q.~A. Wei, S.~Ding, J.~J. Park, R.~Sajnani, A.~Poulenard, S.~Sridhar, and L.~Guibas, ``Lego-net: Learning regular rearrangements of objects in rooms,'' in \emph{Proceedings of the IEEE/CVF Conference on Computer Vision and Pattern Recognition}, 2023, pp. 19\,037--19\,047.

\bibitem{denosingScoreMatching}
P.~Vincent, ``A connection between score matching and denoising autoencoders,'' \emph{Neural Computation}, vol.~23, no.~7, pp. 1661--1674, 2011.

\bibitem{ECCV22HouseKeep}
Y.~Kant, A.~Ramachandran, S.~Yenamandra, I.~Gilitschenski, D.~Batra, A.~Szot, and H.~Agrawal, ``Housekeep: Tidying virtual households using commonsense reasoning,'' \emph{arXiv preprint arXiv:2205.10712}, 2022.

\bibitem{ECCV22TIDEE}
G.~Sarch, Z.~Fang, A.~W. Harley, P.~Schydlo, M.~J. Tarr, S.~Gupta, and K.~Fragkiadaki, ``Tidee: Tidying up novel rooms using visuo-semantic commonsense priors,'' \emph{arXiv preprint arXiv:2207.10761}, 2022.

\bibitem{wu2023tidybot}
J.~Wu, R.~Antonova, A.~Kan, M.~Lepert, A.~Zeng, S.~Song, J.~Bohg, S.~Rusinkiewicz, and T.~Funkhouser, ``Tidybot: Personalized robot assistance with large language models,'' \emph{arXiv preprint arXiv:2305.05658}, 2023.

\bibitem{ramesh2022hierarchical}
A.~Ramesh, P.~Dhariwal, A.~Nichol, C.~Chu, and M.~Chen, ``Hierarchical text-conditional image generation with clip latents,'' \emph{arXiv preprint arXiv:2204.06125}, vol.~1, no.~2, p.~3, 2022.

\bibitem{OpenAI2023GPT4TR}
OpenAI, ``Gpt-4 technical report,'' \emph{ArXiv}, vol. abs/2303.08774, 2023.

\bibitem{radford2021learning}
A.~Radford, J.~W. Kim, C.~Hallacy, A.~Ramesh, G.~Goh, S.~Agarwal, G.~Sastry, A.~Askell, P.~Mishkin, J.~Clark \emph{et~al.}, ``Learning transferable visual models from natural language supervision,'' in \emph{International conference on machine learning}.\hskip 1em plus 0.5em minus 0.4em\relax PMLR, 2021, pp. 8748--8763.

\bibitem{palm-e}
D.~Driess, F.~Xia, M.~S. Sajjadi, C.~Lynch, A.~Chowdhery, B.~Ichter, A.~Wahid, J.~Tompson, Q.~Vuong, T.~Yu \emph{et~al.}, ``Palm-e: An embodied multimodal language model,'' \emph{arXiv preprint arXiv:2303.03378}, 2023.

\bibitem{brohan2022rt}
A.~Brohan, N.~Brown, J.~Carbajal, Y.~Chebotar, J.~Dabis, C.~Finn, K.~Gopalakrishnan, K.~Hausman, A.~Herzog, J.~Hsu \emph{et~al.}, ``Rt-1: Robotics transformer for real-world control at scale,'' \emph{arXiv preprint arXiv:2212.06817}, 2022.

\bibitem{RT-2}
B.~Zitkovich, T.~Yu, S.~Xu, P.~Xu, T.~Xiao, F.~Xia, J.~Wu, P.~Wohlhart, S.~Welker, A.~Wahid \emph{et~al.}, ``Rt-2: Vision-language-action models transfer web knowledge to robotic control,'' in \emph{7th Annual Conference on Robot Learning}, 2023.

\bibitem{huang2023visual}
C.~Huang, O.~Mees, A.~Zeng, and W.~Burgard, ``Visual language maps for robot navigation,'' in \emph{2023 IEEE International Conference on Robotics and Automation (ICRA)}.\hskip 1em plus 0.5em minus 0.4em\relax IEEE, 2023, pp. 10\,608--10\,615.

\bibitem{zhou2023navgpt}
G.~Zhou, Y.~Hong, and Q.~Wu, ``Navgpt: Explicit reasoning in vision-and-language navigation with large language models,'' \emph{arXiv preprint arXiv:2305.16986}, 2023.

\bibitem{huang2023voxposer}
W.~Huang, C.~Wang, R.~Zhang, Y.~Li, J.~Wu, and L.~Fei-Fei, ``Voxposer: Composable 3d value maps for robotic manipulation with language models,'' \emph{arXiv preprint arXiv:2307.05973}, 2023.

\bibitem{lin2023text2motion}
K.~Lin, C.~Agia, T.~Migimatsu, M.~Pavone, and J.~Bohg, ``Text2motion: From natural language instructions to feasible plans,'' \emph{arXiv preprint arXiv:2303.12153}, 2023.

\bibitem{liang2023code}
J.~Liang, W.~Huang, F.~Xia, P.~Xu, K.~Hausman, B.~Ichter, P.~Florence, and A.~Zeng, ``Code as policies: Language model programs for embodied control,'' in \emph{2023 IEEE International Conference on Robotics and Automation (ICRA)}.\hskip 1em plus 0.5em minus 0.4em\relax IEEE, 2023, pp. 9493--9500.

\bibitem{ahn2022can}
M.~Ahn, A.~Brohan, N.~Brown, Y.~Chebotar, O.~Cortes, B.~David, C.~Finn, C.~Fu, K.~Gopalakrishnan, K.~Hausman \emph{et~al.}, ``Do as i can, not as i say: Grounding language in robotic affordances,'' \emph{arXiv preprint arXiv:2204.01691}, 2022.

\bibitem{ha2023scaling}
H.~Ha, P.~Florence, and S.~Song, ``Scaling up and distilling down: Language-guided robot skill acquisition,'' \emph{arXiv preprint arXiv:2307.14535}, 2023.

\bibitem{song2020score}
Y.~Song, J.~Sohl-Dickstein, D.~P. Kingma, A.~Kumar, S.~Ermon, and B.~Poole, ``Score-based generative modeling through stochastic differential equations,'' \emph{arXiv preprint arXiv:2011.13456}, 2020.

\bibitem{dormand1980family}
J.~R. Dormand and P.~J. Prince, ``A family of embedded runge-kutta formulae,'' \emph{Journal of computational and applied mathematics}, vol.~6, no.~1, pp. 19--26, 1980.

\bibitem{zhang2019graph}
X.~Zhang, C.~Xu, X.~Tian, and D.~Tao, ``Graph edge convolutional neural networks for skeleton-based action recognition,'' \emph{IEEE transactions on neural networks and learning systems}, vol.~31, no.~8, pp. 3047--3060, 2019.

\bibitem{cao2021suctionnet}
H.~Cao, H.-S. Fang, W.~Liu, and C.~Lu, ``Suctionnet-1billion: A large-scale benchmark for suction grasping,'' \emph{IEEE Robotics and Automation Letters}, vol.~6, no.~4, pp. 8718--8725, 2021.

\bibitem{MMD}
P.~Achlioptas, O.~Diamanti, I.~Mitliagkas, and L.~Guibas, ``Learning representations and generative models for 3d point clouds,'' in \emph{International Conference on Machine Learning}.\hskip 1em plus 0.5em minus 0.4em\relax PMLR, 2018, pp. 40--49.

\end{thebibliography}
